\documentclass[twoside]{article}

\usepackage{PRIMEarxiv}

\usepackage{booktabs} 
\usepackage{graphicx} 
\usepackage{latexsym} 

\usepackage{amsmath,amssymb,amsfonts} 
\usepackage{algorithm,algpseudocode} 
\usepackage{xcolor} 
\usepackage{enumitem}
\usepackage{url} 

\begin{document}
\title{An effective and efficient green federated learning method for one-layer neural networks}

\author{Oscar Fontenla-Romero, Bertha Guijarro-Berdiñas,
       Elena Hernández-Pereira, Beatriz Pérez-Sánchez \\
    Universidade da Coruña, CITIC \\
    Campus de Elviña, s/n \\
    15071, A Coruña, Spain  \\
    \texttt{\{oscar.fontenla, berta.guijarro, elena.hernandez, beatriz.perezs\}@udc.es}
}

\pagestyle{fancy}
\thispagestyle{empty}
\rhead{ \textit{ }}

\fancyhead[LO]{A green federated learning method}
\fancyhead[RE]{Oscar Fontenla-Romero et al.}

\maketitle

\begin{abstract}
Nowadays, machine learning algorithms continue to grow in complexity and require a substantial amount of computational resources and energy. For these reasons, there is a growing awareness of the development of new green algorithms and distributed AI can contribute to this. 
Federated learning (FL) is one of the most active research lines in machine learning, as it allows the training of collaborative models in a distributed way, an interesting option in many real-world environments, such as the Internet of Things, allowing the use of these models in edge computing devices. In this work, we present a FL method, based on a neural network without hidden layers, capable of generating a global collaborative model in a single training round, unlike traditional FL methods that require multiple rounds for convergence. This allows obtaining an effective and efficient model that simplifies the management of the training process. Moreover, this method preserve data privacy by design, a crucial aspect in current data protection regulations.
We conducted experiments with large datasets and a large number of federated clients. Despite being based on a network model without hidden layers, it maintains in all cases competitive accuracy results compared to more complex state-of-the-art machine learning models. Furthermore, we show that the method performs equally well in both identically and non-identically distributed scenarios. Finally, it is an environmentally friendly algorithm as it allows significant energy savings during the training process compared to its centralized counterpart.
\end{abstract}


\section{Introduction}

We are all conscious of the enormous potential of artificial intelligence (AI) and their impact in our daily lives. However, it comes together with increases of energy and resource footprint, resulting in clear environmental costs. Machine learning algorithms keep growing in complexity, and many state-of-the-art methods continue to emerge, each requiring a substantial amount of computational resources and energy. On one hand, training and adjustment are energy-consuming processes that can last thousands of hours on specialized hardware accelerators and, by themselves, can produce a high carbon footprint \cite{Wu2021}; on the other hand, very often they depend on the use of data centers that are extremely energy consuming \cite{Qiu2021}. This problem is much more pronounced in recent years when training data and the power of AI models have grown exponentially. As an example, the amount of training data on Facebook for use of recommendations has increased by 2.4 in the last two years, reaching the scale of exabytes and, in OpenAI, the compute used in
various large AI training models had been
doubling every 3.4 months since 2012 \cite{Dhar2020}. This is what has been called Red AI. For these reasons, there is an increasing awareness among the various actors (researchers, governments, industry, etc.) of the importance of developing technological solutions that make an efficient use of energy and other resources, thereby promoting long-term technological sustainability and reducing the environmental impact of AI on our planet. These are the goals of what has been called Green AI. This field is attracting more attention and it can be considered as a hot research topic\footnote{As examples, it is a core topic within the European Green Deal Strategy and countries like Spain already have the development of “green algorithms” as a priority in its National AI Strategy.}.

Currently, there are advances that contribute to avoiding high consumption processes in the training of ML models. Among them, we can mention distributed AI, a computing paradigm that bypasses the need to move huge amounts of data and provides the ability to analyze data at the source (edge devices). While most data processing is still taking place at centralized data centers, organizations are discovering the benefits of edge computing. It was predicted that by 2025, 75 percent of data will be created and processed outside a traditional cloud or data center \cite{Gartner} as a result of the use of mobile devices, the Internet of Things (IoT) or machine-generated data, leading to a rethink about where computing should take place. Therefore, decentralized alternatives to data center based on machine learning are emerging. One of the most prominent is Federated Learning (FL).

Federated learning is a machine learning environment in which many clients collaboratively train a model, usually under the control of a central coordinator, keeping the training data decentralized. This paradigm presents several advantages compared to traditional and centralized machine learning among them, (a) highly efficient use of network bandwidth, less information is transmitted; (b) low latency, real time decisions can be made locally at the different clients and (c) privacy, under the assumption that the clients and servers are non malicious, privacy enhances and the probability of threats reduces to a certain extent. The term federated learning was introduced in 2017 by McMahan et al. \cite{McMahan2017} when the authors proposed to solve a learning task by a loose federation of participating devices (clients) that are coordinated by a central server. The underlying idea of training models without collecting raw training data in a single location has proven to be useful in a wide range of practical scenarios. While FL is still not a fully mature technology, it is already being used by millions of users on a daily basis. The algorithm proposed by McMahan et al. has been applied to Google's Gboard \cite{Hard2018} to improve next word prediction models and several studies have explored the use of federated learning in scenarios that deal with privacy sensitive data such as health diagnosis, collaboration across multiple hospitals or government agencies \cite{Brisimi2018,Verma2018}.

In this work, we present a green federated learning method that is privacy-preserving by design. Unlike most FL methods, it allows to obtain the global model in a non-iterative way in a single round of consultation with the clients, thus reducing both training time and communication needs. In addition, in the event of a client connection failure, client's data can be quickly and easily incorporated into the global model as soon as it becomes available and, as will be shown, it works correctly whether or not the data is identically distributed among the clients. All this makes it an very interesting environmentally friendly algorithm that saves energy during the training process compared to its centralized counterpart.  Reducing the energy consumption of FL techniques will have important benefits, such as reducing the operating costs when training models, increasing battery lifetimes of the devices using those models, and also lowering the temperature of chips and other components, thus making faster running times possible.

In this regard, the potential environmental impact of these algorithms is not being addressed in depth by the scientific community. Whilst the carbon footprint for centralized learning has been studied in many previous works \cite{Lacoste2019}\cite{Henderson2022}, in case of FL environment it remains little explored. There are only few works that deal with this topic \cite{Qiu2021}. Savazzi et al. \cite{savazzi2021aframework} propose a novel framework for the analysis of energy and carbon footprints in distributed and federated learning. The proposed framework quantifies both the energy footprints and the carbon equivalent emissions for methods and consensus-based fully decentralized approaches. Inspired by this work, in this paper, we will use measurements to show the electrical consumption of the proposed model. Specifically, instead of using measures of carbon emissions, which are dependent on the electric power generation systems employed in each country, in this research we will use consumption measures in watt hours, since it is a more general measure.

The remainder of this paper is structured as follows. In Section \ref{sec:RelatedWork}, a review of the main works related to federated learning is carried out. Section \ref{sec:Proposed} details the proposed method. Section \ref{sec:experiments} shows the results of the experimental study carried out with large datasets in independent and identically distributed (IID) and non-IID scenarios. Finally, Section \ref{sec:conclusions} presents the conclusions of the work.

\section{Related work}\label{sec:RelatedWork}

In the federated learning environment, models are trained in a distributed scenario, either across a large number of clients with low computational abilities, such as personal devices, or across a relatively small number of very reliable clients, typically organizations such as medical or financial institutions. The first setting is known as cross-device while the second one is known as cross-silo. In any case,  FL training is usually conducted in several rounds of communication, in each of which a fraction of the clients are selected to participate in the learning process. These clients receive the current global model from the coordinator and perform local training with their local data before sending the updated models back to the central server. Finally, the server aggregates these updated models, resulting in a new global model. This process is repeated for a fixed number of rounds.
There exists several aggregation strategies, the most widely used is FedAvg \cite{McMahan2017}. In this approach, the central server aggregates the model by performing a weighted sum of the received parameters based on the number of samples in each local dataset. FedAvg has probed its success but, it has been shown that its performance degrades significantly on heterogeneous data. FedAvg reduces the frequency of communication required performing multiple updates on the local clients, but this fact can lead to overfitting to local data. This phenomenon is called \textit{client-drift} and can lead to slower convergence. 
To solve this problem, Karimireddy et al. present a solution named SCAFFOLD \cite{Karimireddy2020a}. SCAFFOLD uses a mechanism to correct the \textit{client-drift} in its local updates, requiring significantly fewer communication rounds and not being so affected by the heterogeneity of the data. But its main drawback is that it is designed for what is called a cross-silo setting, in which  only a relatively small number of reliable clients participate usually
with large computational ability (e.g. institutions). To solve it, a new solution MIME \cite{Karimireddy2020b} was presented for the cross-device federated learning setting, where the number of clients may be extreme, they may have limited computing resources (e.g. mobiles) and  the communication network may be unreliable. MIME is a framework that adapts centralized algorithms to the federated setting. It overcomes the natural client-heterogeneity in such a setting and reduces communication by naively reducing the number of participating clients per round. MIME uses a combination of control variates and server-level optimizer state (e.g. momentum) at every client-update step to ensure that each local update mimics that of the centralized method run on IID data. Although it has been shown to consistently outperform FedAvg, and largely mitigate the \textit{client-drift} problem, it still exhibits some performance perturbations compared to the centralized base algorithm.

Finally, regarding energy consumption, Alfauri el al \cite{Alfauri2021adistributed} propose an Energy-efficient Consensus Protocol (EECP) for sustainable federated learning. The goal of their protocol is to jointly optimize computation and transmissions by selecting at each round which clients will participate, while the total energy consumption is minimized. Although this protocol is applicable in almost any federated learning scheme, the method is not concerned with energy-optimizing the specific learning algorithm used at each client.

\section{Proposed method}\label{sec:Proposed}

In \cite{Fontenla2023}, a federated method is described that includes strong security measures to prevent privacy breaches by using homomorphic encryption for client-to-client communications. However, under conditions where communications can be assumed to be secure and clients are not malicious, dispensing with this encryption process would yield a more energy efficient method while still achieving minimal privacy by design.

Based on this, we propose in this paper a new federated learning method for one-layer feedforward neural networks (no hidden layers). We will first define the terminology based on a model with a centralized training set and, then, we will describe the federated learning scheme in detail. Let us consider a training dataset defined by an input matrix $\mathbf{X} \in \mathbb{R}^{m\times n}$, being $m$ the number of inputs (including the bias) and $n$ the number of data instances, and a desired output matrix $\mathbf{D} \in \mathbb{R}^{n \times c}$, where $c$ is the number of outputs. In what follows, we consider only one output (i.e., $\mathbf{d} \in \mathbb{R}^{n \times 1}$) to avoid a cumbersome derivation; however, the extension to multiple outputs is straightforward, since the one-layer neural network each output depends only on a set of independent weights. The parameters of the neural network are defined by a weight vector $\mathbf{w} \in  \mathbb{R}^{m \times 1}$ and the output $\mathbf{y}$ of the model can be obtained by the following equation:
\begin{equation}
\mathbf{y} = f(\mathbf{X}^T \mathbf{w})
\end{equation}

\noindent where $f: \mathbb{R} \rightarrow \mathbb{R}$ is the nonlinear activation function at the output neuron.
Our proposal is based on some previous research that, for reasons of clarity and completeness, we will comment briefly in the next subsection.

\subsection{Background}

In neural networks the optimal weights are usually obtained by an iterative process that minimizes a cost function. In the case of supervised learning, there are several alternatives for the cost function and, among them, the mean squared error (MSE) is one of the most used. Normally, the MSE is measured at the output of the network, comparing the real outputs $\mathbf{y}$ and the desired outputs $\mathbf{d}$. However, as proposed in \cite{Fontenla2010}, an alternative option is to minimize the MSE measured \emph{before} the activation function, i.e., between $\mathbf{X}^T \mathbf{w}$ and $\mathbf{\bar{d}} = f^{-1}(\mathbf{d})$. In addition, to avoid overfitting, a regularization term based on the L2 norm can be incorporated. Under these two conditions, the following cost function can be used \cite{Fontenla2021}:

\begin{equation}\label{cost_fun}
J(\mathbf{w}) = \frac{1}{2} \left[\left(\mathbf{F} \left(\mathbf{\bar{d}}  - \mathbf{X}^T\mathbf{w}\right) \right)^T \left(\mathbf{F} \left(\mathbf{\bar{d}}  - \mathbf{X}^T\mathbf{w}\right) \right) + \lambda \textbf{w}^T\textbf{w} \right]
\end{equation}

\noindent where the first summand refers to the MSE calculated before the activation function and the second corresponds to the regularization factor. $\mathbf{F} =  diag(f'(\bar{d_1}), f'(\bar{d_2}),\ldots, f'(\bar{d_n}))$ is a diagonal matrix formed by the derivative of the $f$ function for the components of $\mathbf{\bar{d}}$, whilst the hyperparameter $\lambda$
is a positive scalar that controls the influence of the penalty term and can be set to zero if regularization is not considered necessary.

The cost function in equation (\ref{cost_fun}) has the advantage that, despite using nonlinear activation functions, it is convex and its global minimum can be obtained directly by means of a closed solution, by deriving this function and equating the result to zero. The solution is the $\mathbf{w}$ that satisfies the following system of linear equations \cite{Fontenla2021}:

\begin{equation}\label{sis2}
(\mathbf{X} \mathbf{F} \mathbf{F} \mathbf{X}^T + \lambda \mathbf{I}) \mathbf{w} = \mathbf{X} \mathbf{F} \mathbf{F} \mathbf{\bar{d}}  
\end{equation}

The size of the system of linear equations in equation \ref{sis2} depends on $m$, as the most expensive operation to obtain $\mathbf{w}$ is the calculation of the inverse of the factor that multiplies it, with computational complexity of $O(m^3)$. When $m$ is small, this complexity implies high efficiency, but becomes computationally demanding when the number of inputs is large. 
In order to obtain $\mathbf{w}$ in the most efficient way possible, irrespective of whether the data contains a greater number of samples than variables or vice versa, a transformation of equation (\ref{sis2}) was proposed in \cite{Fontenla2021} using singular value decomposition (SVD). This allows factoring matrix $\mathbf{X}\mathbf{F}$ as $\mathbf{X}\mathbf{F} = \mathbf{U}\mathbf{S}\mathbf{V}^T$ where $\mathbf{U} \in \mathbb{R}^{m \times m}$ and $\mathbf{V} \in \mathbb{R}^{n \times n}$ are orthogonal matrices, and $\mathbf{S} \in \mathbb{R}^{m \times n}$ is a diagonal matrix with $r$ nonzero elements, $r = rank(\mathbf{X}\mathbf{F})\leq min(m, n)$. Using this approach the equation (\ref{sis2}) can be rewritten as follows:

\begin{equation} \label{sis3}
(\mathbf{U} \mathbf{S} \mathbf{V}^T \mathbf{F} \mathbf{X}^T + \lambda\mathbf{I})\mathbf{w} = \mathbf{X} \mathbf{F} \mathbf{F} \mathbf{\bar{d}} 
\end{equation}

\noindent allowing to obtain another closed-form solution \cite{Fontenla2021} to calculate $\mathbf{w}$ equivalent to equation (\ref{sis3}): 

\begin{equation}\label{opt_w}
\mathbf{w} = \mathbf{U} (\mathbf{S} \mathbf{S}^T + \lambda \mathbf{I})^{-1} \mathbf{U}^T \mathbf{X} \mathbf{F} \mathbf{F} \mathbf{\bar{d}} 
\end{equation}

As the number of nonzero elements in the diagonal matrix $\mathbf{S}$ is $r$, the effective dimensions of $\mathbf{U}$ and $\mathbf{S}$ are $m \times r$ and $r \times r$, respectively. This allows to calculate SVD($\mathbf{X}\mathbf{F}$) using an economy-sized decomposition that can be obtained more efficiently, as well as the inverse of $(\mathbf{S} \mathbf{S}^T + \lambda \mathbf{I})$. Under these circumstances, being $r\leq min(m, n)$, the system in equation (\ref{opt_w}) becomes efficient whether $m \gg n$ or $n \gg m$, as demonstrated in \cite{Fontenla2021}.  

\subsection{Federated learning method}

The approach presented in equation (\ref{opt_w}) can only be applied in a centralized learning scenario, where the whole dataset $\mathbf{X}$ is available in a single location, but not in a federated environment where the data is distributed among several clients. When a federated learning scenario is considered, the data is partitioned into $P$ submatrices $\mathbf{X} = [\mathbf{X}_{1}| \mathbf{X}_{2}| \ldots |\mathbf{X}_{P}]$, such that each node (client) contains only one of the submatrices. This is a type of federated learning known as horizontal federated learning, in which it is assumed that all clients have the same variables about the problem but differ in the available samples. In this case, in order to develop our federated proposal, we reformulate the work in \cite{Fontenla2021} taking into account the following considerations:

\begin{itemize}
    \item The first part on the right hand side of equation (\ref{opt_w}), i.e., $\mathbf{U} (\mathbf{S} \mathbf{S}^T + \lambda \mathbf{I})^{-1} \mathbf{U}^T$, involves the matrices $\mathbf{U}$ and $\mathbf{S}$ that must be calculated from the SVD of $\mathbf{X}\mathbf{F}$ in a centralized form. However, Iwen and Ong \cite{Iwen2016} demonstrated that the SVD can be computed in an incremental and distributed way. Their result establishes that given a data matrix $\mathbf{A}$ decomposed into $P$ partitions $\mathbf{A}=[\mathbf{A}_{1}| \mathbf{A}_{2}| \ldots |\mathbf{A}_{P}]$, it has the same singular values $\mathbf{S}$ and left singular vectors $\mathbf{U}$ as the matrix $\mathbf{B}=[\mathbf{U}_{1}\mathbf{S}_{1}| \mathbf{U}_{2}\mathbf{S}_{2} | \ldots | \mathbf{U}_{P}\mathbf{S}_{P}]$, where $\mathbf{U}_{p}\mathbf{S}_{p}\mathbf{V}_{p}=\text{SVD}(\mathbf{A}_{p})$, i.e.:
    \begin{equation}\label{firstTerm}
     \text{SVD}(\mathbf{A})= \text{SVD}([\mathbf{U}_{1}\mathbf{S}_{1}| \mathbf{U}_{2}\mathbf{S}_{2} | \ldots | \mathbf{U}_{P}\mathbf{S}_{P}])
    \end{equation}
    
    This partial SVD merging scheme has been shown to be numerically robust to rounding off errors and corruption of the original data. It is also accurate even when the rank of matrix $\mathbf{A}$ is underestimated or deliberately reduced.
    
    Therefore, by applying this distributed method, we can compute the global economy-sized $\text{SVD}(\mathbf{XF})$ by using the partial results $\mathbf{U}_{p}\mathbf{S}_{p}$ obtained by calculating the $\text{SVD}(\mathbf{X}_p\mathbf{F}_p)$ on every client $p$ in a federated scenario.
    
    \item Also, on the right-hand side of equation (\ref{opt_w}), we have the term $\mathbf{X}\mathbf{F}\mathbf{F}\mathbf{\bar{d}}$, which for simplicity we will call $\mathbf{m} \in \mathbb{R}^{m \times 1}$. This term can also be calculated by partitioning the matrices involved into blocks and applying simple algebraic operations on the resulting submatrices, using the following equation: 
    \begin{eqnarray}\label{incremental_M}
    \mathbf{m} & = & \mathbf{X}\mathbf{F}\mathbf{F}\mathbf{\bar{d}} \\ & = &  [\mathbf{X}_{1}| \mathbf{X}_{2}| \ldots |\mathbf{X}_{P}] \begin{bmatrix} \mathbf{F}_{1} \\ \mathbf{F}_{2} \\ \vdots \\ \mathbf{F}_{P} \end{bmatrix} \begin{bmatrix} \mathbf{F}_{1} \\ \mathbf{F}_{2} \\ \vdots \\ \mathbf{F}_{P} \end{bmatrix} \begin{bmatrix} \mathbf{\bar{d}}_1 \\ \mathbf{\bar{d}}_2 \\ \vdots \\ \mathbf{\bar{d}}_P \end{bmatrix}\\
    & = & \mathbf{X}_1\mathbf{F}_1\mathbf{F}_1\mathbf{\bar{d}}_1 + \ldots + \mathbf{X}_P\mathbf{F}_P\mathbf{F}_P\mathbf{\bar{d}}_P
    \end{eqnarray}
    
    Therefore, $\mathbf{m}$ can be  computed in a distributed manner, through the local information provided by $P$ clients. Suppose that we have a batch of $n_p$ examples $\mathbf{X}_{p}\in \mathbb{R}^{m \times n_p}$ at client $p$ where $\mathbf{m}_p=\mathbf{X}_p\mathbf{F}_p\mathbf{F}_p\mathbf{\bar{d}}_p$ is computed. 
    %
    When new data $\mathbf{X}_{k}$ from another client $k$ become available and $\mathbf{m}_k$ is computed, the new vector $\mathbf{m}_{p|k}$ -- which combines the information provided jointly by $\mathbf{X}_{p}$ and $\mathbf{X}_{k}$ -- can be easily obtained by simply adding to $\mathbf{m}_p$ the term that depends only on the new data block, i.e: 
    \begin{equation}\label{secondTerm}
     \mathbf{m}_{p|k}=\mathbf{m}_p+\mathbf{X}_k\mathbf{F}_k\mathbf{F}_k\mathbf{\bar{d}}_k   
    \end{equation}

\end{itemize}

Taking into account the above considerations, we propose a federated learning model using a client-server architecture. In this architecture $P$ participants, also known as \textit{clients}, who handle data partitions with the same structure in terms of variables, collaboratively train a machine learning model with the help of an aggregation server, also known as \textit{coordinator}. To perform the federated learning at a client $p$, using the local data partition $\mathbf{X}_p$, we only need to compute locally the matrices $\mathbf{U}_p$ and $\mathbf{S}_p$, resulting from the SVD of $\mathbf{X}_p\mathbf{F}_p$, and $\mathbf{m}_p=\mathbf{X}_p\mathbf{F}_p\mathbf{F}_p\mathbf{\bar{d}}_p$. Subsequently, each client sends these partial local results to the coordinator that combines all the information using the result in equation (\ref{firstTerm}) to obtain the global matrices $\mathbf{S}$ and $\mathbf{U}$, and equation (\ref{secondTerm}) to obtain the global $\mathbf{m}$. Then, the weights of the global model learned from all data partitions are calculated using equation (\ref{opt_w}) and distributed among clients to be used to infer on new data.  Since no raw data is transmitted among clients, but only locally computed matrices $\mathbf{U}_p$, $\mathbf{S}_p$ and $\mathbf{m}_p$, the algorithm also guarantees privacy by design. Algorithms \ref{FedHEONN_client} and \ref{FedHEONN_coordinator} contain the pseudocode for both the clients and the coordinator in the proposed federated learning scheme. Also, in the interest of contributing to reproducible research, the Python code is available online\footnote{\url{https://github.com/ofontenla/FedHEONN}.}. 
\begin{algorithm}[h]\nonumber
\textbf{Inputs for a client $p$: } \\
        \phantom{} \hspace{1em} $\mathbf{X}_p \in \mathbb{R}^{m \times n_p}$ \Comment{Local data with $m$ inputs and $n_p$ samples}\\
            \phantom{} \hspace{1em} $\mathbf{d}_p \in \mathbb{R}^{n_p \times 1}$ \Comment{Desired outputs}\\
        \phantom{} \hspace{1em} $f$               \Comment{Nonlinear activation function (invertible)}\\

\textbf{Outputs}: \\
    \phantom{} \hspace{1em} $\mathbf{m}_{p}$  \Comment{Local $\mathbf{m}$ vector computed by client $p$}\\
    \phantom{} \hspace{1em} $\mathbf{U}_p\mathbf{S}_p$ 					    \Comment{Local $\mathbf{U*S}$ matrix computed by client $p$} \\			

\begin{algorithmic}[1]
    \Function {fed\_client}{$\mathbf{X}_p$,$\mathbf{d}_p$,$f$}
		\State $\mathbf{X}_p = [ones(1,n_p); \mathbf{X}_p]$                    \Comment{Bias is added}
        \State $\mathbf{\bar{d}}_p = f^{-1}(\mathbf{d}_p)$                     \Comment{Inverse of the neural function}
        \State $\mathbf{f}_p = f'(\mathbf{\bar{d}}_p)$							\Comment{Derivative of the inverse neural function}
        \State $\mathbf{F}_p = diag(\mathbf{f}_p)$                             \Comment{Diagonal matrix}
    	\State $[\mathbf{U}_p,\mathbf{S}_p,\sim] = SVD(\mathbf{X}_p*\mathbf{F}_p)$ \Comment{Economy size SVD}
		\State $\mathbf{U}_p\mathbf{S}_p= \mathbf{U}_p*diag(\mathbf{S}_p)$ \Comment{Non-zero elements of $\mathbf{S}_p$}
    	\State $\mathbf{m}_p = \mathbf{X}_p*(\mathbf{f}_p.*\mathbf{f}_p.*\mathbf{\bar{d}}_p)$
		\State return $\mathbf{m}_{p}, \mathbf{U}_p\mathbf{S}_p$
    \EndFunction
\end{algorithmic}
\caption{Pseudocode for the client.}
\label{FedHEONN_client}\small
\end{algorithm}
\begin{algorithm}[h]\nonumber
\caption{Pseudocode for the coordinator.}
\label{FedHEONN_coordinator}\small
\textbf{Inputs}: \\
        \phantom{} \hspace{1em} $\mathbf{M}$\_list  \Comment{List containing the $\mathbf{m}$ vectors of the clients}\\
        \phantom{} \hspace{1em} $\mathbf{US}$\_list \Comment{List containing the $\mathbf{US}$ matrices of the clients}\\
        \phantom{} \hspace{1em} $\lambda$           \Comment{Regularization hyperparameter}\\

\textbf{Outputs}: \\
    \phantom{} \hspace{1em} $\mathbf{w} \in \mathbb{R}^{m \times 1}$         \Comment{Weights of the global model to be distributed among clients}\\

\begin{algorithmic}[1]
    \Function {fed\_coordinator}{$\mathbf{M}$\_list, $\mathbf{US}$\_list, $\lambda$}
        \State $\mathbf{m}$  = M\_list[0]              \Comment{Get the first element of the list}
        \State $\mathbf{US}$ = US\_list[0]
        \State for $\mathbf{m}_p$, $\mathbf{US}_p$ in (M\_list[1:], US\_list[1:]): \Comment{From the 2nd}
            \State \hspace{1em} $\mathbf{m}$ = $\mathbf{m}$ + $\mathbf{m}_p$ \Comment{Aggregation of $\mathbf{m}$ vector}
            \State \hspace{1em} $[\mathbf{U},\mathbf{S},\sim] = SVD([\mathbf{US}_p \ | \ \mathbf{US}])$  \Comment{Incremental SVD}
            \State \hspace{1em} $\mathbf{US} = \mathbf{U} * diag(\mathbf{S})$                             \Comment{Update $\mathbf{US}$ to include client $p$}
        \State $\mathbf{w} = \mathbf{U}*inv(\mathbf{S}*\mathbf{S}+\lambda\mathbf{I})*(\mathbf{U}^{T}* \mathbf{m})$   \Comment{Optimal weights}
		\State return $\mathbf{w}$ 
    \EndFunction
\end{algorithmic}
\end{algorithm}

It is worth noting that this federated learning model, by calculating the optimal weights analytically and not iteratively, allows the training process to be performed on clients in a single step and requires only one round of aggregation at the coordinator. This contrasts with the vast majority of the methods proposed in the state of the art which, for a given dataset, usually require several rounds of partial calculations at the clients and aggregation at the coordinator until the global model is obtained.
  
\begin{figure}[htbp]
\centerline{\includegraphics[width=8.5cm]{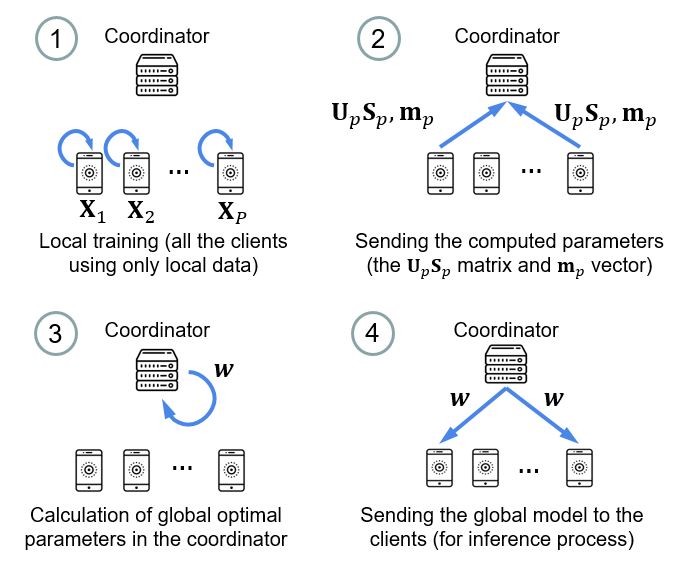}}
\caption{Summary of the main steps for the training process of the proposed model. This process will be repeated each time new data arrives to the clients.} 
\label{fig:FedAlgorithm}
\end{figure}

 As summary, Figure \ref{fig:FedAlgorithm} contains the main steps of the proposed method using a set of clients and a coordinator. Besides, it should be noted that, the coordinator algorithm was described so that it receives the list of matrices and vectors computed by all the clients and adds all of them sequentially. However, this is not a requirement and the coordinator could add clients at different stages if any of the clients were temporarily unavailable. To do this, it would start from the aggregate model with $n$ clients and, incrementally, the client $n+1$ would be added. This also allows the addition of new clients dynamically, since it would not be necessary to retrain the old clients to add the information of the new one in the coordinator.

\section{Experiments} \label{sec:experiments}
In this section, we evaluate the behavior of the proposed method and compare its performance with other results available in the state of the art.
\subsection{Experimental setup}

We evaluated the effectiveness of the proposed method by applying it to big datasets and large number of clients (up to 20,000) in a federated environment. For reference, we also compare the results with the same algorithm but centrally trained, that is, in a single client containing all data. Table \ref{tab:Datasets} shows the classification datasets used and their main characteristics. They are all benchmark and public datasets that are available on the UCI dataset repository \cite{Dua2017}. The dataset called HIGGSx4 was artificially created by replicating four times the data from the Higgs dataset. This experiment was proposed because, although it is not relevant in terms of accuracy, it is very interesting to analyze some other aspects of the behavior of the proposed model in an even more extreme scenario in terms of the number of data.

\begin{table}
\begin{center}
{\caption{Characteristics of the datasets.}\label{tab:Datasets}}
\begin{tabular}{lccc}
\hline
\rule{0pt}{12pt}

Dataset & Samples & Attributes & Classes\\
\hline
\\[-6pt]
SUSY & 5,000,000  & 18 & 2\\
HEPMASS & 10,500,000  & 28 & 2\\
HIGGS &  11,000,000 & 28 & 2\\
HIGGSx4 & 44,000,000 & 28 & 2\\
\hline
\\[-6pt]
\end{tabular}
\end{center}
\end{table}

For all the experiments, the dataset was divided in such a way that 70\% of the data was used for the training process of the model while the remaining 30\% was used for testing. The experiments were repeated three times and the results provided in this study are always the mean value for each of the metrics. 

In all the experiments, a regularization hyperparameter ($\lambda$) equals to $1 \times 10^{-3}$ and logistic activation functions for all the neurons were used.

In order to simulate the federated scenario, the data was distributed among all the clients so that each one had approximately the same number of data in their local database. Table \ref{tab:SizeLocalData} shows the size of the centralized training set and also the approximate size of the local training datasets employed in each of the clients for the most extreme federated scenario (20,000 clients).

\begin{table}
\begin{center}
\caption{Size of training datasets for each experiment in the scenario with the largest number of clients (20,000).}
\label{tab:SizeLocalData}
\begin{tabular}{lcc}
\toprule
& Samples in the & Samples in the\\
Dataset & global training set & local training sets\\
\midrule
SUSY & 3,500,000 & 175 \\
HEPMASS & 7,350,000 & 368 \\
HIGGS & 7,700,000   & 385 \\
HIGGSx4 & 30,800,000 & 1,540\\
\bottomrule
\end{tabular}
\end{center}
\end{table}

Since we do not have 20,000 devices to run the clients on different machines, the federated environment was simulated on the same computer. Thus, the clients of the federated model, and also the centralized model, were executed on a computer with a Intel Core i7-10700 2.90GHz processor with 32GB of RAM.

To check the performance of the model, the following metrics were used:

\begin{itemize}
    \item Test accuracy: percentage of correctly predicted data points out of all the data points in the test set.
    \item Training time: in a real federated environment, all the clients run in parallel on different devices, therefore, the total training time in this environment will be considered as the time required by the slowest client (it would be the last one to send the information to the coordinator) plus the time required by the coordinator to aggregate the information of all the clients.
    \item Sum of CPU time: it is the sum of the individual training times of all the clients plus the time of the coordinator. This CPU time is an important reference as global energy consumption required by the federated system.
    \item Watts per hour (Wh): depending on the watts consumed by the device and the CPU time spent in each one of them for training, the consumption can be calculated in terms of watt per hour. In this case, since all the clients use the same type of device, the calculation is simple since it would be the result of multiplying the watts by the \textit{sum of CPU time} (in seconds) divided by 3,600.
\end{itemize}

Finally, since the data heterogeneity, which may exist between clients and/or due to data imbalance at each client, is one of the main challenges in federated learning scenarios, we performed tests using identically distributed and non-identically distributed data. The results are shown independently in the following sections. 

\subsection{Results for IID scenarios}

In order to simulate the IID scenario, the data were shuffled and distributed randomly among all the clients so that the same proportion of data from all classes is preserved at every client as in the original set. Each client performs a training process with the assigned local data and later sends the information to the coordinator to be added to create the global collaborative model.

\begin{figure}
\centerline{\includegraphics[width=12.5cm]{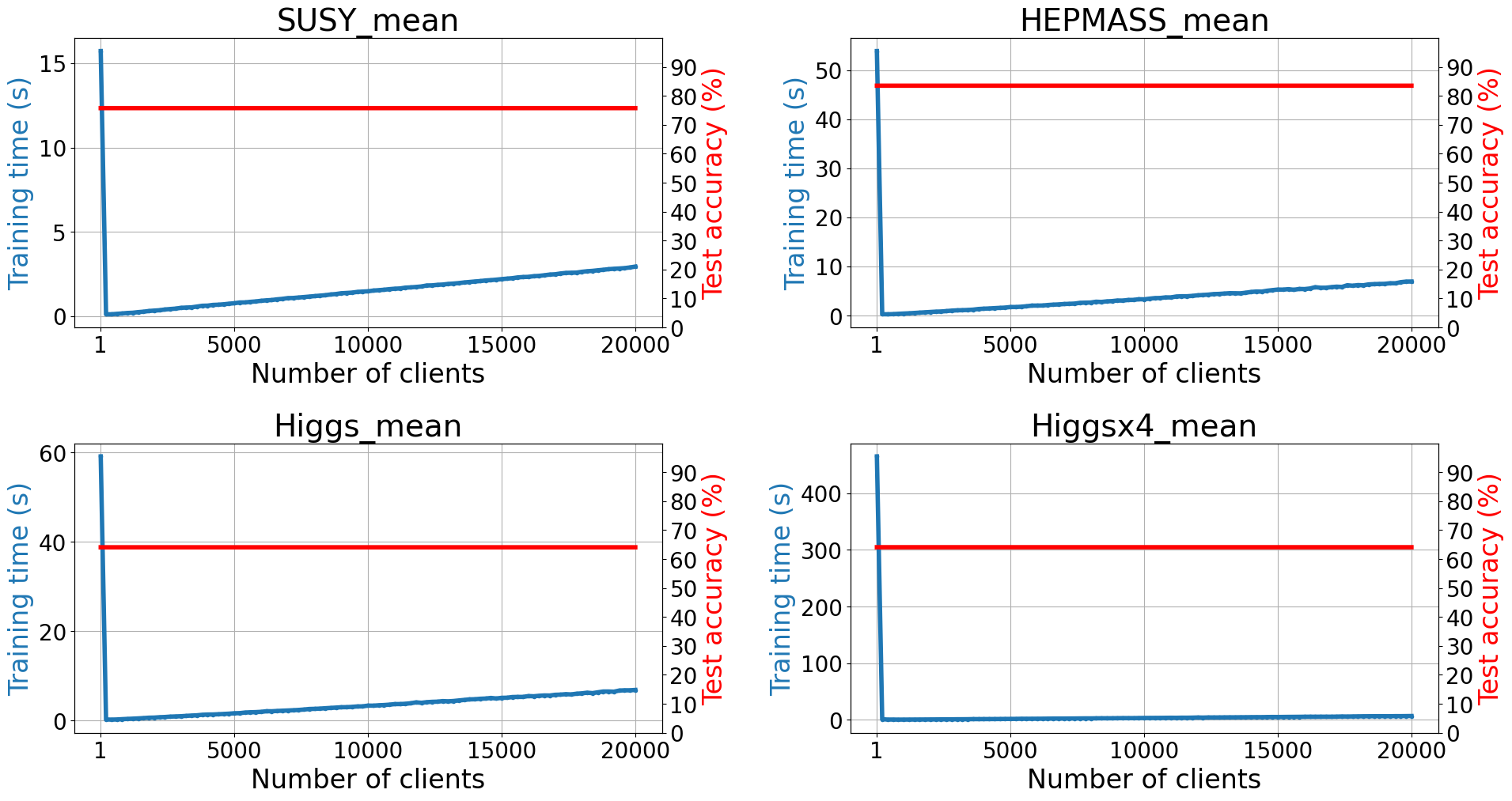}}
\caption{Training time and accuracy varying the number of clients in the IID federated scenario. The results are the mean of the three experiments for each dataset.}
\label{fig:IID_TrainingTime}
\end{figure}

Figure \ref{fig:IID_TrainingTime} contains the results of the training time and accuracy for the four datasets, varying the number of clients from 1 to 20,000, with steps of 200. The first value of the curves (value one of the x-axis) corresponds to the centralized training of the model. The figure contains two vertical axes. The vertical axis located on the left refers to the training time (blue curve) while the vertical axis located on the right represents the accuracy obtained for each scenario (red curve). Both curves indicate the behaviour of the model in each of the client configurations.

As can be seen, the accuracy value does not vary regardless of the number of clients employed, and is identical to that obtained in the centralized scenario. This allows us to state that the performance of the method does not degrade even if the number of clients increases significantly.
Regarding the training time, it can be seen that the federated model maintains a much lower time than the centralized one. Even if the number of clients is increased, and although there is a slight increase in the curves in this case, this increase is very small. 

In short, the model allows a significant increase in the number of clients for large datasets without losing effectiveness or efficiency in its behaviour.

\begin{figure}[htbp]
\centerline{\includegraphics[width=12.5cm]{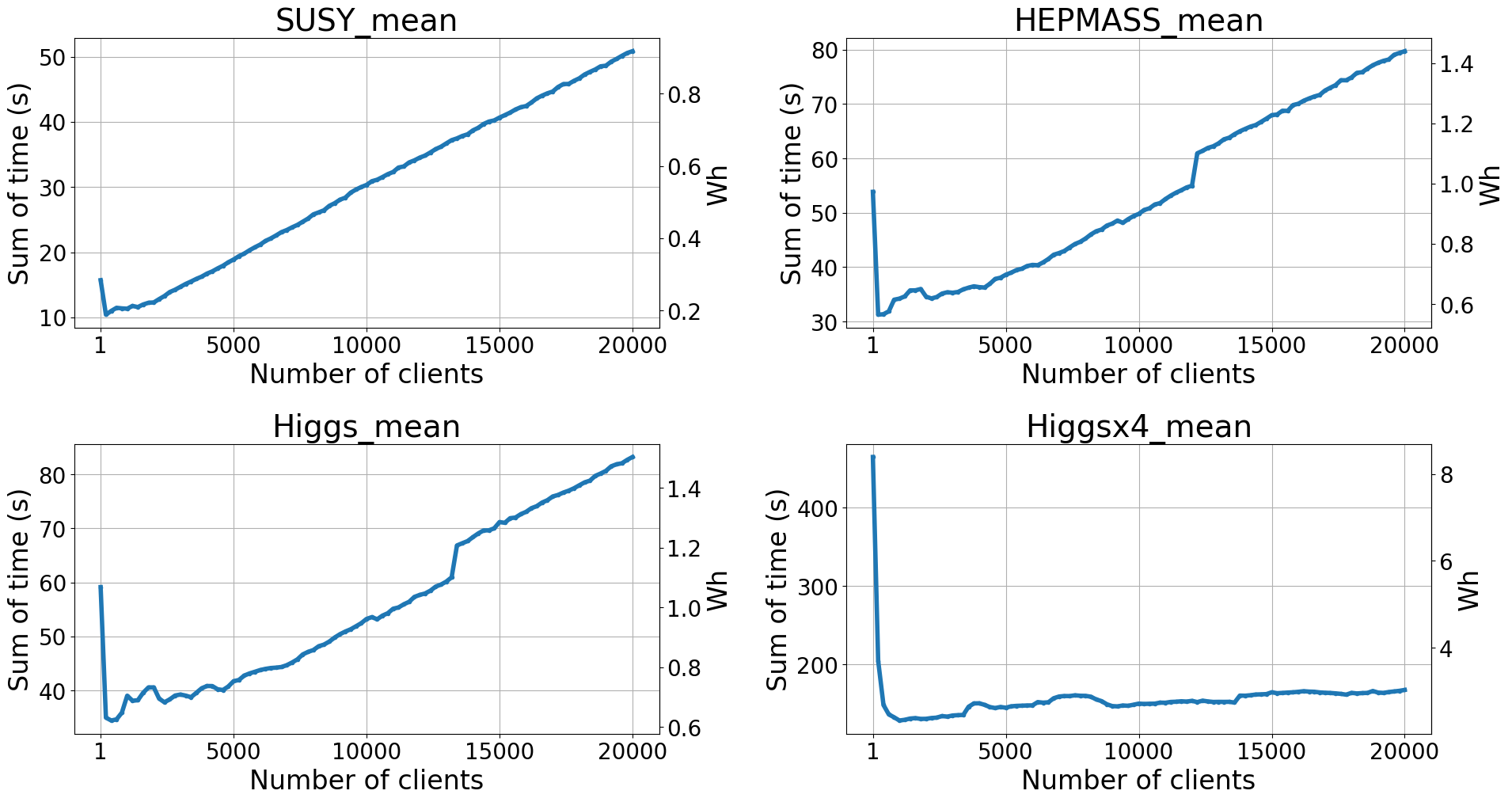}}
\caption{Sum of CPU time and watts per hour consumed in the training process varying the number of clients in the IID federated scenario. The results are the mean of the three experiments for each dataset.}
\label{fig:IID_SumCPUTime}
\end{figure}

In addition, Figure \ref{fig:IID_SumCPUTime} shows, for all the datasets, the \textit{sum of CPU time} and \textit{Watts per hour} consumed varying the number of clients. Again, value 1 of the x- axis corresponds to the centralized training scenario. The main vertical axes, on the left, show the sum of time in seconds and the secondary axes shows the corresponding Watts per hour. The curve is unique for both vertical axes.

As can be seen, the number of clients required to exceed the consumption of a centralized option is directly proportional to the volume of data to be handled. Logically, the centralized option increases its consumption as the size of the training dataset grows. The distributed option generates a lower consumption initially, but there may come a time when the number of devices is so large that it demands more energy than the centralized option. Despite this, in the experiments conducted, in the smallest set (SUSY) the number of devices supported without exceeding the consumption of the centralized option is around 4000, while for the largest set  (Higgsx4), the demand generated by the 20,000 devices used is well below the demand of the centralized version of the algorithm.

\subsection{Results for non-IID scenarios}

To create the non-IID scenario, we first sort all data by class label and, then, they were sequentially distributed, following the established order, among the clients based on the data size assigned to each of them (in this case, the same for all). This is a pathological non-IID partition of the data, as the vast majority of the clients will only have examples of one class.

\begin{figure}[htbp]
\centerline{\includegraphics[width=12.5cm]{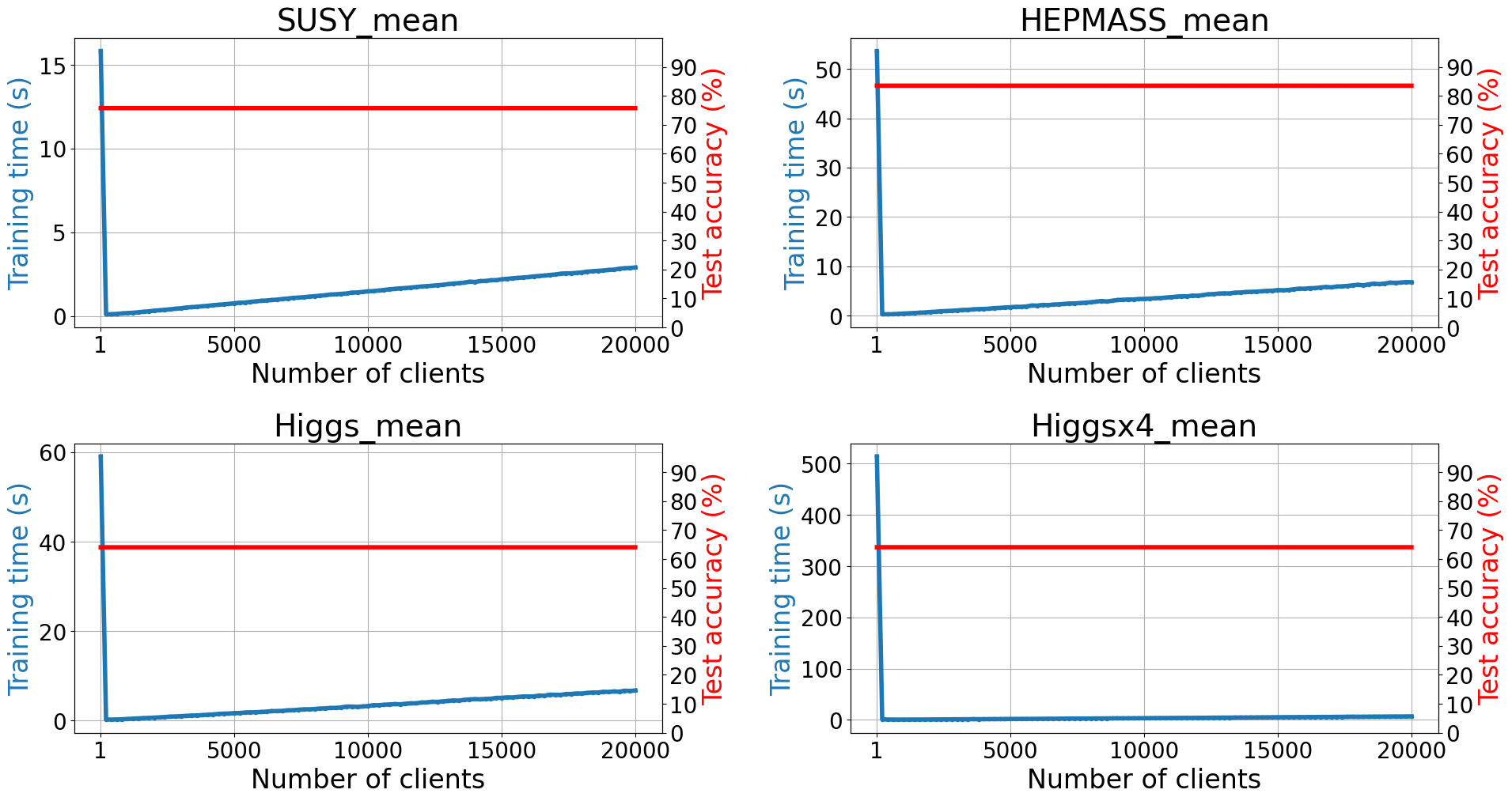}}
\caption{Training time and accuracy varying the number of clients in the non-IID federated scenario. The results are the mean of the three experiments for each dataset.}
\label{fig:NonIID_TrainingTime}
\end{figure}

Figure \ref{fig:NonIID_TrainingTime} contains the results of the training time and accuracy for all the datasets in this scenario. The first value of the curves (value one of the x-axis) is again the one that corresponds to the centralized training of the model. As can be seen, one more time the accuracy is maintained regardless of the number of clients employed and, moreover, it is exactly the same as in the IID scenario, emphasizing the usefulness of client cooperation in a federated environment.

\begin{figure}[htbp]
\centerline{\includegraphics[width=12.5cm]{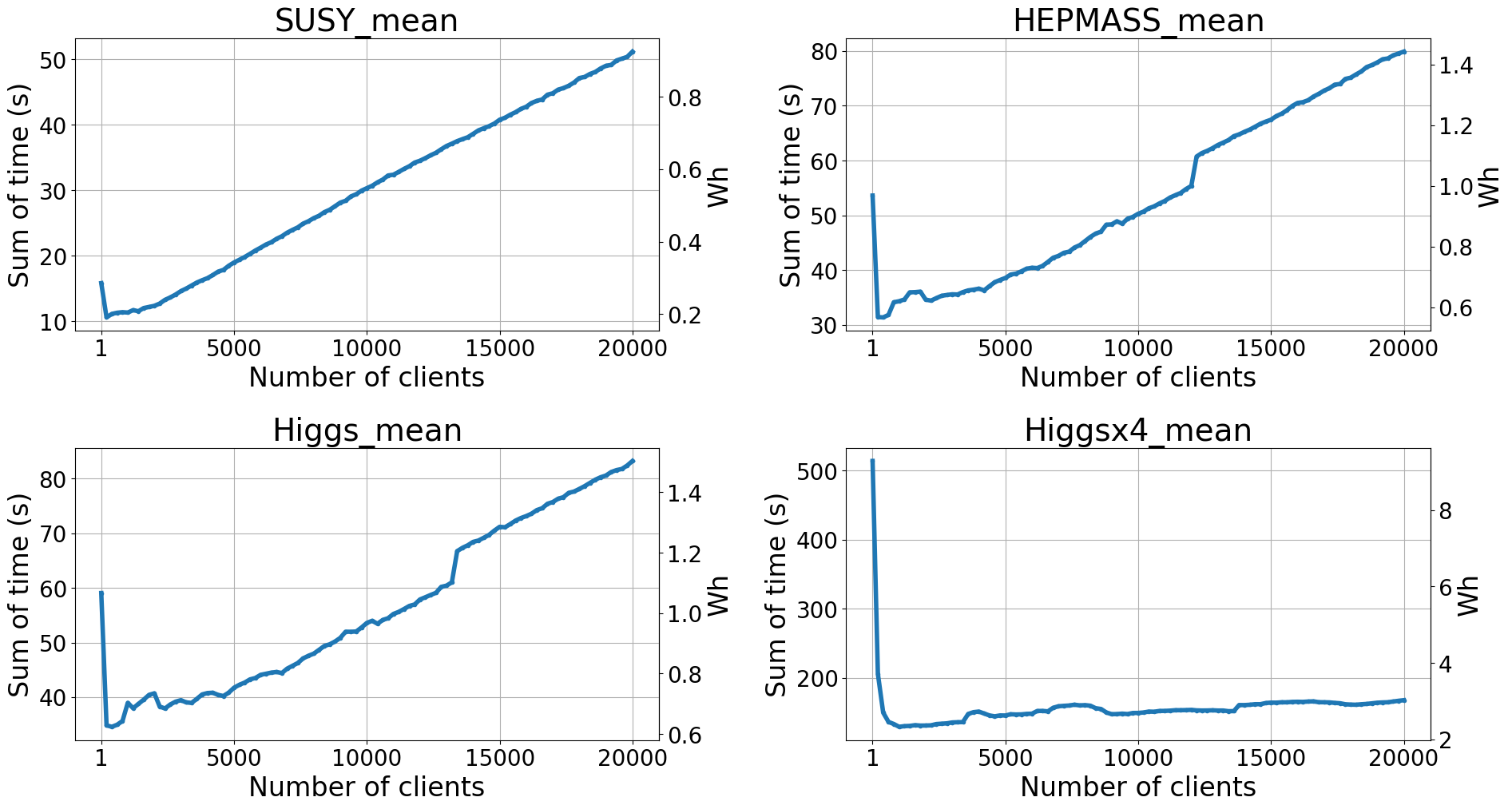}}
\caption{Sum of CPU time and watts per hour consumed in the training process varying the number of clients in the non-IID federated scenario. The results are the mean of the three experiments for each dataset.}
\label{fig:NonIID_SumCPUTime}
\end{figure}

Figure \ref{fig:NonIID_SumCPUTime} shows, for all the datasets, the sum of CPU time and Watts per hour consumed varying the number of clients in this non-IID scenario. Comparing with the equivalent figure of the IID scenario (Figure \ref{fig:IID_SumCPUTime}), it can be seen that the results are very similar, as it would be expected. For all these reasons, it can be affirmed that the method does not present learning problems in non-IID scenarios, which on the other hand are the most common federated environments.

\subsection{Accuracy comparison}

To finish the experimental part, a comparative study of the accuracy results obtained by the model against some other state-of-the-art machine learning models is included. Table \ref{tab:AccuracyComparison} contains the quantitative results of the proposed model, for each dataset, compared to the main ones found in the scientific literature. The first column of the table contains the name of the method and the bibliographical reference from which the result was obtained. Cells with a hyphen indicate that the article mentioned does not include results for that dataset. To facilitate comparison, the last two rows of the table include the average and median value obtained by all the models analyzed, without including the proposed method.

As can be seen, the proposed method, despite being based on a network model without hidden layers, allows to obtain very competitive results even against more complex machine learning models that use all the data in a centralized way, achieving better results in many cases. In addition, comparing with the average value obtained by other models, it can be observed that the proposed method presents a higher performance. Considering the median, it achieves slightly better results in 2 of the 3 sets. All this, with the added advantages that it can be used in a federated learning environment (IID or non-IID) with very low energy consumption to process large datasets.

\begin{table}
\begin{center}
\caption{Accuracy (\%) results of the proposed method compared with those previously published by other authors.}
\label{tab:AccuracyComparison}
\begin{tabular}{lccc}
\toprule
Model & HIGGS & SUSY & HEPMASS \\
\midrule
\textbf{Proposed method}  &	\textbf{64.05}	 & \textbf{75.76}	 & \textbf{83.50} \\ 
Scalable PANFIS Merging	\cite{Pratama2019} & 63.66 & 	76.70	 & 83.47 \\
Scalable PANFIS Voting \cite{Pratama2019} & 63.70 & 	76.22	 & 84.18 \\
Scalable PANFIS AL Merging \cite{Pratama2019} & 	63.72	 & 76.79	 & 83.45 \\
Scalable PANFIS AL Voting \cite{Pratama2019} & 	63.92	 & 76.20 & 	84.15 \\
Spark.Kmeans \cite{Pratama2019} & 	48.34	 & 50.04	 & 50.66 \\
Spark.GLM \cite{Pratama2019} & 	63.51	 & 75.01 & 	83.40 \\
Spark.GBT \cite{Pratama2019} & 	59.49 & 	75.11 & 	81.83 \\
Spark.RF \cite{Pratama2019} & 	59.65 & 	76.81 & 	82.43 \\
PDMS \cite{Laila2022} & 	63.00	 & -	 & 83.67 \\
MMD-D \cite{Liu2020} & 	57.90 & 	- & 	- \\
Logistic Regression \cite{Azhari2020a} & 	64.21 & 	- & 	- \\
Decision tree \cite{Azhari2020a} & 	63.57	 & - & 	- \\
Random Forest \cite{Azhari2020a} & 	67.64 & 	-	 & - \\
Gradient Boosted Tree \cite{Azhari2020a} & 	70.62 & 	- & 	- \\
PANFIS MapReduce \cite{Pratama2018} & 	63.48 & 	76.80 & 	83.35 \\
PANFIS \cite{Pratama2018} & 	63.94 & 	75.42	 & 83.32 \\
eTS \cite{Pratama2018} & 	64.69 & 	77.05	 & 82.32 \\
Simpl\_eTS \cite{Pratama2018} & 	60.17	 & 70.93 & 	81.22 \\
Logistic Regression \cite{Azhari2020b} & 	- & 	78.84 & 	- \\
Random Forest \cite{Azhari2020b} & 	-	 & 77.40 & 	- \\
Decision tree \cite{Azhari2020b} & 	- & 	75.46 & 	- \\
Gradient Boosted Tree \cite{Azhari2020b} & 	- & 	79.30 & 	- \\
MapReduce MRAC \cite{Bechini2016} & 62.96 & 	74.57 & 	- \\
Random Forest \cite{JuezGil2021} & 67.67 & 	77.67 & 	82.21 \\
PCARDE \cite{JuezGil2021} & 	58.33	 & 72.64 & 	81.33 \\
Rotation Forest	\cite{JuezGil2021} & 68.80 & 	78.59 & 	84.44 \\
\midrule
\textbf{Average}  & \textbf{62.86}	& \textbf{74.88} & \textbf{80.96} \\
\textbf{Median}  & \textbf{63.62}	& \textbf{76.46} & \textbf{83.34} \\
\bottomrule
\end{tabular}
\end{center}
\end{table}

\section{Conclusions} \label{sec:conclusions}

In this work, we have presented a novel federated machine learning method, based on one-layer neural networks. It has been shown that for large datasets the proposed method is a very efficient federated proposal that also allows a significant reduction in energy consumption, and as a consequence in carbon emissions, compared to its centralized counterpart. This makes it an algorithm very much in line with green artificial intelligence and well suited for use in learning on edge environments where training processes generally cannot consume large resources. In addition, it also has some other advantages compared to current federated learning methods:

\begin{itemize}
    \item In order to obtain the optimal weights of the network, only a single round of training is necessary, involving all the devices. The vast majority of current methods in the literature use multiple rounds of training in which part of the clients collaborate. This allows for a faster and more robust learning process against attacks that attempt to violate data privacy, by having to share less information over the communication network.
    \item It is privacy preserving by design, as no raw data is transmitted nor can be recovery from the interchanged data.
    \item It has been shown that the method obtains equivalent solutions whether the environment is IID or non-IID. This is an important advantage since most federated learning models do not have exactly the same solution and performance in IID environments as in non-IID ones. Moreover, its accuracy  is similar to the one obtained in a centralized setting.
\end{itemize}

Finally, it should be noted that the method has some limitations that must be taken into account. As it is designed for neural network without hidden layers, it has much less representation power than other deeper networks. Despite this, in some scenarios, for example in IoT environments where low computing power devices are used, this may be the only alternative for clients to be trained on edge devices with low computing performance, such as Raspberry Pi, Jetson Nano, etc.

As future work, we consider the possibility of designing more powerful algorithms for more complex models, such as using the proposed method as a building block for more efficient deeper models.

\section*{Acknowledgments}
This work has been supported by the National Plan for Scientific and Technical Research and Innovation of the Spanish Government (Grants PID2019-109238GB-C2 and PID2021-128045OA-I00); and by the Xunta de Galicia (ED431C 2022/44) with the European Union ERDF funds. CITIC, as a Research Center of the University System of Galicia, is funded by Conseller\'ia de Educaci\'on,  Universidade e Formaci\'on Profesional of the Xunta de Galicia through the European Regional Development Fund (ERDF) and the Secretar\'ia  Xeral de Universidades (Ref. ED431G 2019/01).

\bibliographystyle{unsrt}  
\bibliography{biblio} 

\end{document}